\documentclass[10pt,twocolumn,letterpaper]{article}

\usepackage{iccv}
\usepackage{times}
\usepackage{epsfig}
\usepackage{graphicx}
\usepackage{amsmath}
\usepackage{amssymb}
\usepackage{mathrsfs}

\usepackage{array}
\usepackage{booktabs}
\usepackage{subcaption,siunitx}

\newcolumntype{M}[1]{>{\centering\let\newline\\\arraybackslash\hspace{0pt}}m{#1}}
\newcolumntype{N}[1]{>{\centering\let\newline\\\arraybackslash\hspace{0pt}\bfseries}m{#1}}

\newcolumntype{K}[1]{>{\centering\arraybackslash}p{#1}}

\usepackage[accsupp]{axessibility}
\usepackage[pagebackref=true,breaklinks=true,letterpaper=true,colorlinks,bookmarks=false]{hyperref}

\iccvfinalcopy 

\def\iccvPaperID{4} 

\ificcvfinal\fi

\begin{document}

\title{Toward Affective XAI: Facial Affect Analysis for Understanding\\ Explainable Human-AI Interactions}

\author{Luke Guerdan, Alex Raymond, and Hatice Gunes\\
Department of Computer Science and Technology\\
University of Cambridge, United Kingdom\\
{\tt\small\{lg619, alex.raymond, hatice.gunes\}@cl.cam.ac.uk}

}

\maketitle
\ificcvfinal\thispagestyle{empty}\fi

\begin{abstract}
As machine learning approaches are increasingly used to augment human decision-making, eXplainable Artificial Intelligence (XAI) research has explored methods for communicating system behavior to humans. However, these approaches often fail to account for the affective responses of humans as they interact with explanations. Facial affect analysis, which examines human facial expressions of emotions, is one promising lens for understanding how users engage with explanations. Therefore, in this work, we aim to (1) identify which facial affect features are pronounced when people interact with XAI interfaces, and (2) develop a multitask feature embedding for linking facial affect signals with participants' use of explanations. Our analyses and results show that the occurrence and values of facial AU1 and AU4, and Arousal are heightened when participants fail to use explanations effectively. This suggests that facial affect analysis should be incorporated into XAI to personalize explanations to individuals’ interaction styles and to adapt explanations based on the difficulty of the task performed.

\end{abstract}

\section{Introduction}

Machine learning and artificial intelligence (AI) techniques are being adopted to support human decision-making in domains such as medicine \cite{hamet2017artificial}, business \cite{jarrahi2018artificial}, and education \cite{holstein2021designing}. This accelerating collaboration between humans and AI systems has spurred recent research investigating topics such as communication flow between humans and AI-systems \cite{liang2019implicit}, methods for improving collaboration among teams of humans and agents \cite{bansal2019beyond}, and human attitudes towards AI collaborators \cite{cai2019hello}. An overarching requirement for these agendas is the need to provide humans with meaningful representations of AI decision-making. 

As a result, eXplainable Artificial Intelligence (XAI) methods have been developed to provide humans with greater transparency as they interact with AI systems \cite{vilone2020explainable}. For example, XAI methods have been developed to provide decision makers with salient features used to make a prediction, as well as estimates of model confidence \cite{van2020interpretable}. While approaches such as these can improve human decision-making and increase trust in AI systems \cite{weld2019challenge}, many existing XAI methods remain centered on technical aspects of explanations and fail to account for cognitive-emotional processes \cite{samek2017explainable}. Modeling the affective and cognitive states of users as they interact with AI systems is critical, as these factors have been shown to introduce bias in the decision-making process and skew user perceptions of human-agent collaboration \cite{george2016affect, weld2019challenge}. 

Previous work has incorporated affective information into the explanation generation process \cite{kaptein2017role, 8272592, PeCoX}. For example, \cite{PeCoX} formulates explanations as originating from an agent with goals, beliefs, and emotional processes. While this research lays important groundwork for emotionally-aware XAI, it does not incorporate human affective \textit{responses} to explanations. By modeling human affective responses, it may be possible to (1) personalize explanations to individuals' specific interaction styles, and (2) adapt explanations based on the difficulty of the task the user is performing. 

Therefore, this work investigates human facial affective responses to an XAI interface in the context of a multi-agent navigation task and provides the following contributions.
Firstly, we aim to establish which facial affect features are relevant in an explanation-aware context. We examine three standard representations (valence/arousal, categorical, and facial AUs) used for facial affect analysis, and three alternate facial affect prediction models (OpenFace \cite{baltrusaitis2018openface} and two state-of-the-art multitask methods \cite{deng2020multitask}). We find that (1) these methods provide consistent affect predictions (along a chosen subset), and (2) \textit{objective} facial affect predictions are correlated with \textit{subjective} user perceptions of competence, task difficulty, agency, and explanation utility.
Secondly, we aim to connect facial affect features with participants' use of explanations. We examine how explanations and task difficulty interact to influence facial affect signals. Specifically, we compare participants' facial affect \textit{with} and \textit{without} explanations across three difficulty levels. We then examine whether predictions differ during key game events (e.g., post-collision). We find that several affect predictions (1) vary as a function of task difficulty, and (2) vary during difficult game events. Finally, using principal component analysis (PCA), we develop a multitask embedding that leverages a combination of affect predictions to capture explanation-related affective signals. We find two components that capture aspects of participant interactions, including one that may be heightened when participants do not understand explanations.

\section{Related Work} \label{section:background}

We now introduce three alternate representations for encoding facial affect. We then describe a recently proposed multitask method that we use to compute facial affect estimates for all three representations simultaneously.   

\subsection{Models of Affect}
Three methods for encoding affective information are: (1) a dimensional model  \cite{russell1980circumplex}, (2) a categorical model \cite{ekman1971constants}, and (3) a Facial Action Coding System (FACS) model \cite{wathan2015equifacs}. A dimensional model encodes emotions in continuous space, most often a two-dimensional valence and arousal space \cite{russell1980circumplex}. Valence represents whether an emotion is positive or negative, while arousal represents emotion intensity \cite{nicolaou2011continuous}. The categorical emotion model encodes emotions as a discrete set of options: happiness, anger, sadness, surprise, disgust, and fear. The FACS coding system encodes human facial expressions in terms of the facial action units (AUs) that compose the overall expression \cite{wathan2015equifacs}. Each AU represents a facial behavior that is generated with an anatomically distinct facial muscle group \cite{cohn2007observer}.
There are also works that have investigated the connection between AUs and the basic emotion categories \cite{fiorentini2012identification, kohler2004differences, mehu2012reliable}. Specifically, AU1 (Inner brow Raiser) has been associated with fear and surprise \cite{fiorentini2012identification}, AU2 (Outer Brow Raiser) has been associated with fear and happiness \cite{kohler2004differences}, and AU4 (Brow Lower) has been linked with fear, disgust, and anger \cite{fiorentini2012identification, kohler2004differences}. AU25 (Lips Part) has also been associated with emotions such as surprise. Figure \ref{fig:hyp2} provides example images of these  \cite{facs}.

\subsubsection{Componential Models} \label{section:componential}
%
The appraisal theory of emotion proposes that emotional components are caused by an appraisal, in which the individual determines whether a stimulus (1) matches their goals and expectations, (2) is easy or difficult to control, and (3) can be overcome with the appropriate coping mechanisms \cite{scherer1999appraisal}. An appraisal model of emotion is particularly relevant in an \textit{emotionally-aware XAI} context \cite{kaptein2017role}. This is because the user is leveraging explanations of AI behavior to achieve an \textit{overarching goal}, and has \textit{agency} to decide how to incorporate information provided from the model. Under this framework, explanations of a complex system can be framed as a coping tool that either supports the individual in achieving their goal, or a hindrance that impedes their performance. To discover which emotion features are most relevant in this \textit{emotionally-aware XAI} context, we leverage a multitask emotion recognition approach.

\subsection{Multitask Emotion Recognition} \label{section:multitask}

Whereas many video emotion recognition models are trained to predict AUs, emotional expressions, or valence-arousal in isolation, recent work proposes predicting all three tasks simultaneously using a multitask model \cite{deng2020multitask, kollias2019expression}. A challenge of this approach is that there are few emotion recognition datasets that have annotations for all three tasks \cite{kollias2018aff}. To solve this problem, a recently proposed method leverages knowledge distillation to train a model with missing labels \cite{deng2020multitask}. First, each of the teacher models learns one of the tasks independently based on task-specific ground-truth labels. Then, an ensemble of student models learn all tasks simultaneously using a combination of ground-truth labels and soft labels generated by the three teacher models. 

Given a sample of data $(X, Y)=\{x^{i}, y^{i}\}_{i=1}^{3}$ annotated with one labels for one of the tasks $i$ and a teacher network parameterized by $\theta_t$, the teacher loss function can be defined as: \vspace*{-3mm}

\begin{equation} \label{eq:teacher}
    \mathscr{F}_{\mathrm{t}}\left(X, Y, \theta_{\mathrm{t}}\right)=\sum_{i=1}^{3} \mathscr{L}^{(i)}\left(y^{(i)}, f_{\theta_{\mathrm{t}}}^{(i)}(x^{(i)})\right)
\end{equation}

where $\mathscr{L}^{(i)}$ is the supervision loss function for task $i$. The loss function for the student model combines the supervision loss defined by the teacher function (as in e.q. \ref{eq:teacher}) above with a \textit{distillation loss function} $\mathscr{H}^{(i)}$ that quantifies the error between the teacher and student predictions. Given a student network parameterized by $\theta_s$, the student loss function can be defined as:

\begin{align}
\begin{split}
 \mathscr{F}_{\mathrm{s}}\left(X, Y, \theta_{\mathrm{t}}, \theta_{\mathrm{s}}\right)=\sum_{i=1}^{3}\left\{\lambda \cdot \mathscr{L}^{(i)}\left(y^{(i)}, f_{\theta_{\mathrm{s}}}^{(i)}(x^{(i)})\right)\right. \\
 \enspace +(1-\lambda) \cdot \mathscr{H}^{(i)}\left(f_{\theta_{\mathrm{t}}}^{(i)}(x^{(i)}), f_{\theta_{\mathrm{s}}}^{(i)}(x^{(i)})\right) \\
 \enspace +\sum_{j \neq i} \left. \mathscr{H}^{(j)}\left(f_{\theta_{\mathrm{t}}}^{(j)}(x^{(j)}), f_{\theta_{\mathrm{s}}}^{(j)}(x^{(j)})\right)\right\}
\end{split}
\end{align}

Where $\lambda$ is a constant that controls how much the loss considers the ground-truth label, $\mathscr{H}^{(i)}$ encodes distillation loss within tasks, and $\mathscr{H}^{(j)}$ encodes distillation loss between tasks. In this way, the student model learns to perform all three tasks simultaneously by learning a generalized multitask representation. Additionally, the authors also propose training an ensemble of student models in conjunction to sample a broader multitask representation space. The underlying model structure (layers, activations, etc.) of $\theta_t$ and $\theta_s$ are kept the same to support knowledge transfer \cite{deng2020multitask}. 

This was evaluated with two architecture variants: a Multitask CNN with a ResNet50 backbone architecture, and a Multitask CNN+RNN that uses the same ResNet50 backbone architecture in conjunction with gated recurrent units (GRUs). Whereas the CNN is used for frame-by-frame video feature extraction, the RNN extends the model to capture temporal dynamics within the video. Figure \ref{fig:pipeline} provides a diagram with the details of both of these architectures. 

This model is trained using a combination of the Aff-Wild2 dataset \cite{kollias2018aff}, which contains subsets of data annotated for each of the three tasks, and separate datasets annotated with labels for one of the three tasks in isolation \cite{mavadati2013disfa, kossaifi2017afew, zhang2018facial}. This approach provided state-of-the-art results in the FG-2020 ABAW Competition \cite{kollias2020analysing}.

\section{Methodology} \label{section:methodology}

In this section, we describe the dataset used in this work and outline the expected findings before describing techniques used in this work.

\subsection{Dataset}

An existing human-agent interaction dataset was used for emotion analysis. The dataset consists of event logs and video recordings of 31 participants as they complete a multi-agent navigation task in an easy, medium, or hard difficulty level with and without explanations of agent behavior \cite{raymond2020culture}. During the task, participants played a Busy Barracks (BB) computer game, in which they navigated across a planar grid towards a goal position.

Participants were given a fixed fuel budget and asked to avoid colliding with a set of agents that move with their own desired trajectories. Specifically, during each move, participants were tasked with either (a) moving north, south, east, or west or (b) remaining in a fixed position at a cost of one fuel unit. Participants were penalized 5 fuel units for colliding with an agent, 1 fuel unit for each move or wait action, and 1 fuel unit for every 10 seconds of game clock time. Therefore, participants were incentivized to move directly towards the goal while minimizing agent collisions. 

Participants were assigned to one of three difficulty levels: $L_{E}$, $L_{M}$, or $L_{H}$. Each level corresponds to a set of rules governing agent interactions and behavior, which become increasingly complex as difficulty increases. Participants played their assigned difficulty level \textit{with} augmented explanations of agent behavior ($X$ condition) and \textit{without} augmented explanations of agent behavior ($N$ condition), where the ordering was randomized among participants. Additionally, participants completed an experience survey at the end of each session. Questions were adapted from a standard Game Experience Questionnaire (GEQ) \cite{ijsselsteijn2013game}, and were clustered into Competence, Affect, Challenge, and game-specific questions. Answers were given on a 5-point Likert scale. Additional details regarding the study design and dataset are provided in the original publication \cite{raymond2020culture}.

\subsection{Affective Behavior Analysis} \label{section:hypothesis}

The authors of the path-deconfliction dataset reported that scores increase and collisions decrease when explanations are provided in $L_{H}$. Conversely, collisions increase and scores decrease when participants are provided with explanations in $L_{E}$ \cite{raymond2020culture}. In terms of the appraisal theory of emotion, in $L_{H}$, explanations may afford participants with an \textit{appropriate coping mechanism} for reaching their \textit{overarching goal} of avoiding collisions. Therefore, participant performance improves. In $L_{E}$, explanations may interfere with participants' goals and make it more difficult for them to reach their objective. However, it remains unclear precisely how participant facial affect is related to their ability to use explanations successfully. To gain an understanding of this relationship, we formulate the following hypotheses:\\ 

\noindent\textbf{Explanation-Related}
\begin{itemize}
    \item[\textbf{H1.A}] In $L_H$, predicted affect will be more positive when participants are provided explanations.
    \item[\textbf{H1.B}] In $L_E$, predicted affect will be more negative when participants are provided explanations.
\end{itemize}
\textbf{Event-Related}
\begin{itemize}
    \item[\textbf{H2.A}] Participants will display more negative affect during collision events.
    \item[\textbf{H2.B}] Participants will display more positive affect during success events.
\end{itemize}  

We define a collision event as the 10 seconds of recording immediately after a participant collides with an agent. Conversely, we define a success event as the 10 seconds before a participant reaches the goal position. Given the directive ``to reach the goal while avoiding collisions with agents", we posit that these events may be emotionally salient for participants. Further, these hypotheses operationalize positive affect as (1) high valence scores, (2) happiness expression predictions, and (3) increased frequency of AU6 (Cheek Riser) and AU12 (Lip Corner Puller) activations. Conversely, negative affect is operationalized as (1) low valence scores, (2) anger, disgust, and sadness expression predictions, and (3) increased AU4 (Brow Lower), AU1 (Inner Brow Raiser) activations. Facial AUs were linked with positive and negative emotions based on componential analysis literature discussed in Section \ref{section:componential}.

\subsection{Affect Prediction}

Of the 35 participants in the human-agent path deconfliction dataset, 31 agreed to provide video recordings as part of data collection. Six additional participants were excluded from video analysis because of video formatting issues (3 subjects), or because camera orientation occluded a frontal view of the face in the majority of recording frames (3 subjects). This resulted in a final dataset of 25 participants, where 10 participated in $L_{E}$, 8 participated in $L_{M}$, and 7 participated in $L_{H}$. 

\begin{figure}[!th]
  \includegraphics[width=\linewidth]{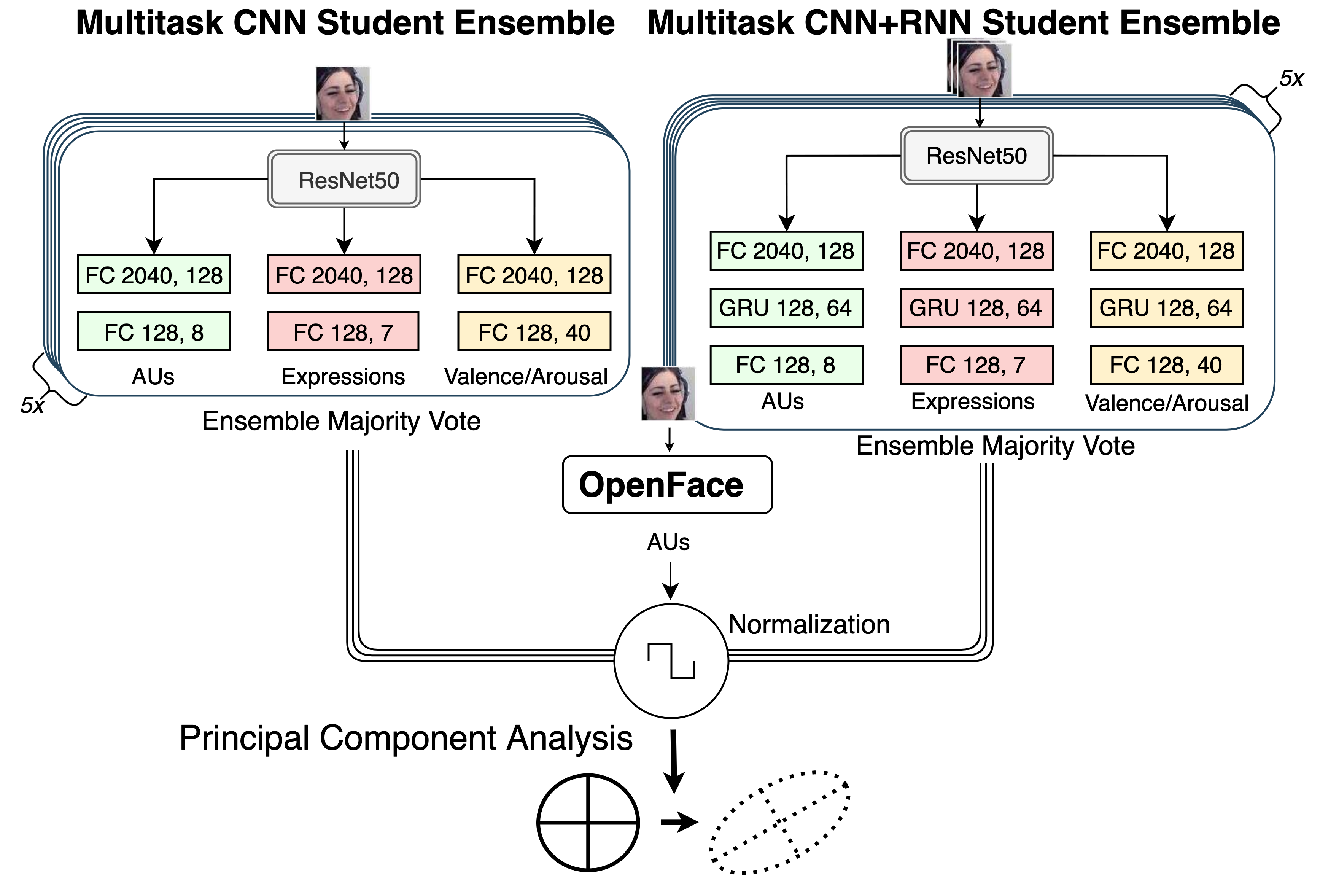}
  \caption{ Emotion recognition and component processing pipeline, including multitask architectures from \cite{deng2020multitask}.} \label{fig:pipeline}

\end{figure}

Three techniques were used to extract AU, valence-arousal, and emotion expression predictions. The first approach used was the OpenFace Facial AU Recognition module, a commonly-used facial behavioral analysis toolkit \cite{baltrusaitis2018openface, 7284869}. 
The other two emotion processing approaches used were the Multitask CNN and Multitask CNN+RNN discussed in Section \ref{section:multitask}. After face detection and alignment using MTCNN \cite{zhang2016joint}, a pretrained Multitask CNN and a pretrained Multitask CNN+RNN provided by \cite{deng2020multitask} were used to predict results for all three task types using an ensemble of 5 student models each. The Multitask CNN+RNN with a sequence length of 8 (as opposed to the others of 16 and 32) was used due to large memory requirements and processing times required by longer sequence lengths. Figure \ref{fig:pipeline} provides an overview of the full extraction pipeline.

\vspace*{-3mm}
\subsection{Reliability Analysis} \label{section:reliability}

Because the path deconfliction dataset has not been annotated with affect labels, it is not possible to compare each facial affect prediction result to a ground-truth baseline. Therefore, a reliability analysis was conducted to measure the agreement between the Multitask CNN, Multitask CNN+RNN, and OpenFace predictions. Inter-rater reliability analysis is commonly conducted in affective computing studies to (1) establish the level of agreement between multiple emotion annotations, or (2) establish the degree to which a new annotation agrees with an established ground truth \cite{siegert2014inter}. In this work, we wish to establish the agreement between several annotations without a ground truth. 

To provide a granular comparison of the three approaches, a pairwise metric comparing each set of metrics is favored. Although Pearson's  ${\displaystyle r}$ has been used extensively \cite{siegert2014inter}, it is not appropriate for comparing nominally-encoded variables \cite{chaturvedi2015evaluation}. Therefore, separate reliability measures were used for the categorically encoded AU and expression predictions, and the continuously encoded valence/arousal predictions. 

\subsubsection{Reliability Measures}
Cohen's $\kappa$ is appropriate for establishing agreement between two categorical ratings. It assumes each annotation is nominally encoded, and occupies a set of mutually-exclusive categories \cite{mchugh2012interrater}. The $\kappa$ statistic produces a value from -1 to 1, where agreement may be 0 = chance-level; 0.10 - 0.20 = slight; 0.21 - 0.40 = fair; 0.41 - 0.60 = moderate; 0.61-0.80 = substantial; 0.81 - 0.99 = near-perfect; and 1.00 = perfect \cite{mchugh2012interrater}.

Similarly, Lin's Concordance Correlation Coefficient (CCC) is well-suited for the continuous valence/arousal comparison \cite{lawrence1989concordance}. The CCC is commonly used to evaluate inter-rater agreement on valence/arousal annotations in affective computing contexts \cite{siegert2014inter}. Like the Pearson correlation, $CCC$ ranges from -1 to 1, with 1 indicating perfect agreement, -1 indicating a perfect inverse relationship, and 0 indicating no relationship \cite{lawrence1989concordance}. The $CCC$ was selected over the standard Pearson correlation because it accounts for chance levels of agreement between variables. 

\vspace*{-2mm}
\subsubsection{Evaluation Procedure}

An Action Unit reliability analysis was conducted on the subset of AUs that are available through all three extraction approaches: AU1, AU2, AU4, AU4, AU12, AU15, AU20, and AU25. To maintain consistency with the binary-encoded facial AUs extracted via the multitask methods, only the binary presence/absence AUs extracted by OpenFace were included in this analysis. Because OpenFace does not include categorical emotion predictions or valence/arousal scores, the reliability analysis of these features was only conducted for the Multitask CNN / Multitask CNN+RNN method combination. Each comparison was conducted on the subset of frames that have annotation scores for each of the three approaches (i.e.\ no methods dropped a frame due to face recognition failure). 

\begin{table}
    \begin{subtable}{1\linewidth}
    \sisetup{table-format=-1.2}   
    \centering
    \scalebox{0.85}{
        \begin{tabular}{K{2.35cm}|K{.48cm}K{.44cm}K{.44cm}K{.44cm}K{.55cm}K{.55cm}K{.55cm}K{.55cm}}
            \toprule
                Cohen's $\kappa$  & AU1 &  AU2 &  AU4 &  AU6 &  AU12 &  AU15 &  AU20 &  AU25 \\
                \midrule
                CNN-RNN & 0.13 & 0.13 & 0.82 & 0.06 & 0.80 & 0.01 & 0.37 &  0.61 \\
                CNN-OpenFace & 0.01 & 0.11 & 0.16 & 0.05 &  0.35 &  0.00 & 0.01 &  0.17 \\
                RNN-OpenFace & 0.20 & 0.34 & 0.22 & 0.21 & 0.38 & 0.00 & 0.04 & 0.27\\
                \bottomrule
            \end{tabular}
         }
    \end{subtable}
    
    \begin{subtable}{1\linewidth}
        \sisetup{table-format=-1.2}   
        \centering
         \scalebox{0.85}{
                \begin{tabular}{K{1.7cm}|K{.7cm}K{.58cm}K{.75cm}K{.32cm}K{1.13cm}K{.82cm}K{.78cm}}
                \toprule
                 Cohen's $\kappa$ &   Neutral &  Anger & Disgust & Fear & Happiness &   Sadness &  Surprise \\
                \midrule
                CNN-RNN & 0.30 &  0.46 &     0.0 & 0.0 & 0.37 &  0.16 &  0.10 \\
                \bottomrule
            \end{tabular}
         }
    \end{subtable}
    \caption{Cohen's $\kappa$ for facial AUs (top) and emotion categories (bottom).} \label{table:agreement}
    \vspace*{-4mm}
\end{table}

\subsection{Componential Modeling}

Given AU, emotion expression, and valence/arousal predictions derived from the proposed pipeline, we now wish to use these multitask emotion features to link participant affect with their use of explanations. In particular, we wish to examine their \textit{appraisals} of the game situation as they use explanations to achieve their goal. One potential approach would be to predict events such as collisions using supervised learning. However, such an approach may not be ideal because the dataset is small, there are few collision events, and collision events are imbalanced among participants. We therefore favor an unsupervised learning approach that attempts to find \textit{performance and explanation related affect embeddings} within the multitask feature representation. By examining (1) whether extracted components are functionally related to participant performance, and (2) to what degree each affect predictor contributes to the component, it may be possible to find an interpretable representation of participant affect as they interact with explanations. 

An embedding describes a representation of information that occupies a lower-dimension (often continuous) space and offers interpretable representations of a high-dimension feature space. Embedding approaches have been proposed for affective computing tasks in previous work \cite{han2019emobed, mohammadi2019towards}. One simple approach for finding an interpretable embedding is principal component analysis (PCA), which finds components that explain a maximum amount of variance in the data \cite{wold1987principal}. 
The full set of all extracted features is transformed with PCA in order to establish (1) whether component loadings use the same features across the methods, and (2) certain multitask features are associated with one another in component space. 

To perform PCA, affect predictions were combined from all tasks derived from the Multitask CNN and Multitask RNN models, i.e., expression categories, valence/arousal values, and binary AUs. These were combined with continuous and binary AU predictions from OpenFace. Second, the set of features were normalized within-subjects using the mean and variance of each feature for each subject across both the $N$ and $X$ conditions. Finally, a PCA transform was performed on the normalized set of 85 features, and the resulting component vectors were used to transform the affect recognition results into component space. Analysis of these components in terms of their loadings and temporal activations is provided in the following section.   

\section{Results} \label{section:results}

We now present results from our analyses of reliability and componential modeling, and connect these with user performance and self-reported experiences.

\subsection{Reliability Analysis}

Table \ref{table:agreement} shows Cohen's $\kappa$ for facial AU and emotion expression features. For AU predictions, results indicate that Cohen's $\kappa$ is generally highest between Multitask CNN and Multitask CNN-RNN predictions, and the lowest between the Multitask CNN and OpenFace predictions. Additionally, the AU4, AU12, and AU25 agreement levels are higher across methods than agreement levels for AU1, AU2, AU6, AU5, and AU20. Among the set of AU4, AU12, and AU25, the CNN-RNN agreement levels are classified in the ``substantial" to ``nearly-perfect" agreement range, while the CNN-OpenFace and RNN-OpenFace agreement levels generally fall under ``fair". 

For categorical emotion predictions, results indicate that agreement between the Multitask CNN and Multitask RNN is ``chance-level" for sadness and surprise, ``fair" for happiness and neutral, and ``substantial" for anger. The Multitask RNN did not predict disgust or fear in any frames, which resulted in a chance-level agreement of 0. Additionally, the $CCC$ of Multitask CNN and Multitask CNN+RNN predictions was .77 and .84 for valence and arousal, respectively. This suggests high prediction reliability  \cite{lawrence1989concordance}. 

An earlier comparison 
found that the Multitask CNN+RNN out-performs the Multitask CNN in most contexts \cite{deng2020multitask}. While no work has directly compared the reliability of multitask models with OpenFace AU predictions, 
OpenFace \cite{ baltrusaitis2018openface} is 
known to provide reliable results in an array of environmental conditions \cite{calvo2010affect}.

\subsection{Affective Behavior Hypotheses}\label{sec:hyps}

The affect predictions used to evaluate \textbf{H1} and \textbf{H2} were informed by the results of the reliability analysis. Because Cohen's $\kappa$ was low between Multitask CNN and OpenFace AU predictions, only the Multitask RNN+CNN and OpenFace predictions were used to compare AU results. AUs were only included in the analysis if Cohen's $\kappa$ between the Multitask CNN+RNN and OpenFace fell into a ``fair" range or higher. This resulted in AU1, AU2, AU4, AU6, AU12, and AU25 being included in analysis. The final AU predictions used to evaluate the hypotheses were determined by averaging the predictions of the Multitask CNN+RNN with the binary predictions derived from OpenFace. 

\begin{figure}[!th]
  \includegraphics[trim={3mm 3mm 3mm 19mm},clip,width=\linewidth]{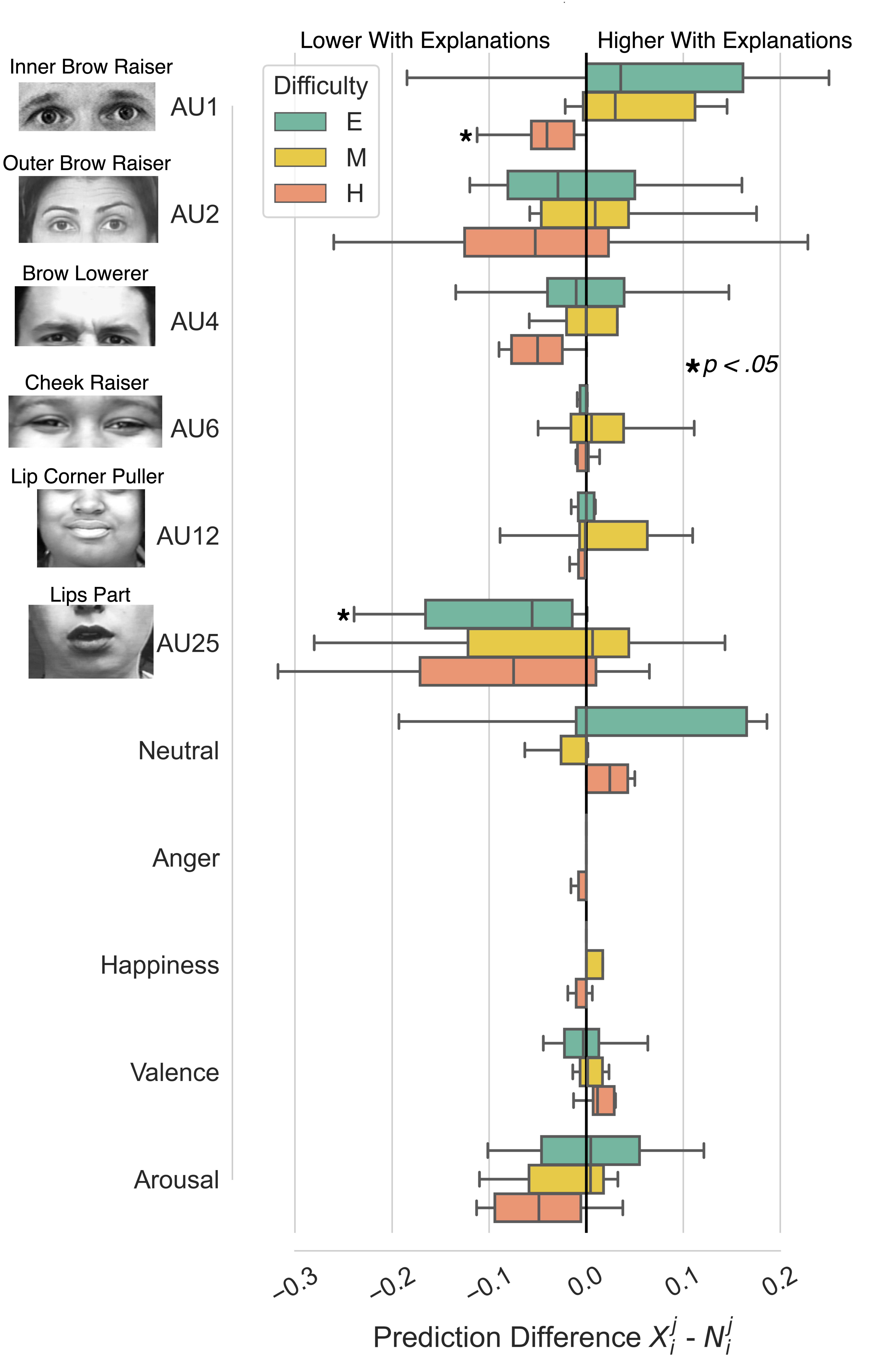}
  \caption{$D_i^j$ across $L_{E}$, $L_{M}$, and $L_{H}$. Facial AU examples from public FACS archive \cite{facs}.} \label{fig:hyp2}
  \vspace*{-7mm}
\end{figure}

Though previous evaluations indicate that the Multitask CNN+RNN produces more accurate categorical and valence/arousal predictions \cite{deng2020multitask}, it is unknown whether this holds for this dataset because of the lack of ground-truth labels. Therefore, predictions from both the Multitask CNN+RNN and Multitask+CNN were used in analysis. Emotion categories were only included in the analysis if Cohen's $\kappa$ between the Multitask CNN and Multitask CNN+RNN fell in the ``fair" range or higher. As a result, sadness, surprise, fear, and disgust predictions were excluded. Because valence and arousal demonstrated a high $CCC$, both were included in analysis. The final categorical and valence/arousal predictions used to evaluate the hypotheses were determined by averaging the predictions from the Multitask CNN+RNN and Multitask CNN. 

\textbf{H1.A} states that affect should be more positive in $L_{H}$ with explanations, while \textbf{H1.B} states that affect should be more negative in $L_{E}$ with explanations. These hypotheses are evaluated by comparing participant affect between the $X$ and $N$ conditions. Specifically, by comparing the difference in the average prediction for feature $i$ between the $X$ and $N$ conditions for each participant $j$ such that $D_i^j = X_i^j - N_i^j$, it is possible to tell whether explanations increased the frequency of the feature. If $D_i^j$ \textit{is negative, this suggests that the feature is less frequently predicted} in the $X$ condition for participant $j$. Conversely, if $D_i^j$ \textit{is positive, this suggests that the feature is more frequently predicted} with explanations for participant $j$. Figure \ref{fig:hyp2} shows this metric for each feature across $L_{E}$, $L_{M}$, and $L_{H}$.

Results support \textbf{H1.A}. In the $L_{H}$ condition, explanations reduce the presence of AU1 and AU4 (linked with fear, surprise, and anger), lower arousal, and increase the proportion of frames rated as neutral. Conversely, in $L_{E}$, explanations increase the frequency of AU1, but results do not conclusively support the claim that participants have more negative affect when explanations are presented in the easy condition. Thus \textbf{H1.B} is not supported. We also conduct a significance test for each feature. A Shapiro–Wilk test indicated samples from $D_i^j$ are not normally distributed for several features. Therefore, because sample sizes are also small, with 10, 8, and 7 participants in $L_E$, $L_M$, $L_H$ respectively, we use a non-parametric 1-Sample Wilcoxon Signed-Rank test. We test whether the median of $D_i^j$ is zero for each feature $i$ and difficulty setting $L_E$, $L_M$, $L_H$. Here, $H_0$: the median of $D_i$ is zero, and $H_1$: the median of $D_i$ is nonzero (two-sided). With $\alpha=0.05$, $H_0$ was rejected for AU1 in $L_H$ ($Z=0.0, p=0.027$) and AU25 in $L_E$ ($Z=1.0, p=0.010$).  

\textbf{H2} holds that predicted affect will be more positive during success events and less positive during collision events. To evaluate this, a collision event difference was computed for feature $i$ such that $DC_i = C_i - C_i'$, where $C_i$ are frames during a collision event, and $C_i'$ are frames outside of a collision event. The success event difference was computed using the same method such that $DS_i = S_i - S_i'$, where $S_i$ and  $S_i'$ are frames inside and outside of success event windows, respectively. Positive values of $DC$ and $DS$ indicate that a feature is more prominent during the event, while negative values indicate that it is less prominent. Results of this analysis are provided in Table \ref{table:hyp3}. They show that valence is slightly lower during collision events, and slightly higher during success events. Results also show that arousal is higher during collision and success events. AU1, AU2, and Neutral are predicted less frequently during collision events. Thus, \textbf{H2.A} and \textbf{H2.B} are supported. 

Taken together, these results indicate that (1) several facial affect predictors (AU1, AU25, Neutral, and Arousal especially) differ in response to explanations of agent behavior, and (2) several others (AU1, AU2, and Neutral in particular) differ in response to key game-play events.

\begin{table}[!th]
    \centering
    \scalebox{0.85}{
    \begin{tabular}{M{1.8cm}|M{1.8cm}|M{1.8cm}}
    \toprule
    \textbf{Feature} & \textbf{Collision Event} &  \textbf{Success Event}  \\
    \midrule
    AU1       &      -0.07 &       -0.03 \\
    AU2       &      -0.11 &        0.00 \\
    AU4       &       0.02 &       -0.04 \\
    AU6       &       0.02 &        0.02 \\
    AU12      &       0.00 &        0.04 \\
    AU25      &       0.02 &       -0.02 \\
    Neutral   &      -0.09 &       -0.02 \\
    Anger     &       0.01 &        0.01 \\
    Happiness &       0.02 &        0.02 \\
    Valence   &      -0.01 &        0.02 \\
    Arousal   &       0.05 &        0.03 \\

    \bottomrule
    \end{tabular}
    }
    \caption{Difference between predicted emotion features from frames occurring during collision events ($DC$) and frames occurring during success events ($DS$).}\label{table:hyp3}
    \vspace*{-4mm}
\end{table}

\begin{table*}[!th]
    \centering
    \scalebox{0.8}{
    \begin{tabular}{N{1.5cm}|M{1.1cm}|M{.9cm}|M{.9cm}|M{.9cm}|M{.9cm}|M{.9cm}|M{.9cm}|M{.9cm}|M{.9cm}|M{.9cm}|M{.9cm}|M{.9cm}|M{.9cm}|M{.9cm}}
    \toprule\hline
     Comp. 1 & AU45 & AU2 & AU1 & AU9 & AU2 & AU1 & AU45 & AU15 & AU2 & AU25 & AU25 & AU25 & AU1 & AU20  \\ 
     \hline
     \textbf{Method} &  OFP &  OFP &  OFP &  OFP &  CNN & OFI & OFI & OFI & OFI &  CNN &  OFP &  RNN &  CNN & OFI\\
     \hline
    \textbf{Loading} &   .39 &               .29 &               .27 &              -.26 &           .23 &                .22 &                .22 &               -.21 &                .21 &           .21 &               .18 &           .16 &           .16 &               -.14\\ 
    \hline\toprule
         \textbf{Comp. 2}  &   Neut. &             AU4 &             AU4 &                AU4 &                AU7 &         Neut. &                AU17 &                 AU5 &                 AU2 &                 AU7 &                 AU14 &                 AU20 &            AU25 &                AU10  \\
     \hline
     \textbf{Method} & CNN &  RNN &  CNN &  OFP &  OFP &  RNN &  OFP & OFI & OFI & OFI  & OFI & OFI &  RNN &  OFP\\
     \hline
    \textbf{Loading} &   -.41 &           .32 &           .30 &               .30 &               .25 &          -.23 &               .22 &               -.18 &               -.18 &                .17 &               -.16 &               -.15 &          -.14 &               .12  \\
    \hline\bottomrule
    \end{tabular} 
    }
    \caption{Top 14 loadings for Component 1 (top) and Component 2 (bottom). CNN denotes Multitask CNN, RNN denotes Multitask RNN, OFP denotes Open Face presence, OFI denotes Open Face intensity.} \label{table:loadings}
    \vspace*{-5mm}
\end{table*}

\begin{figure*}
  \includegraphics[trim={1mm 2mm 3mm 2mm},clip,width=\linewidth]{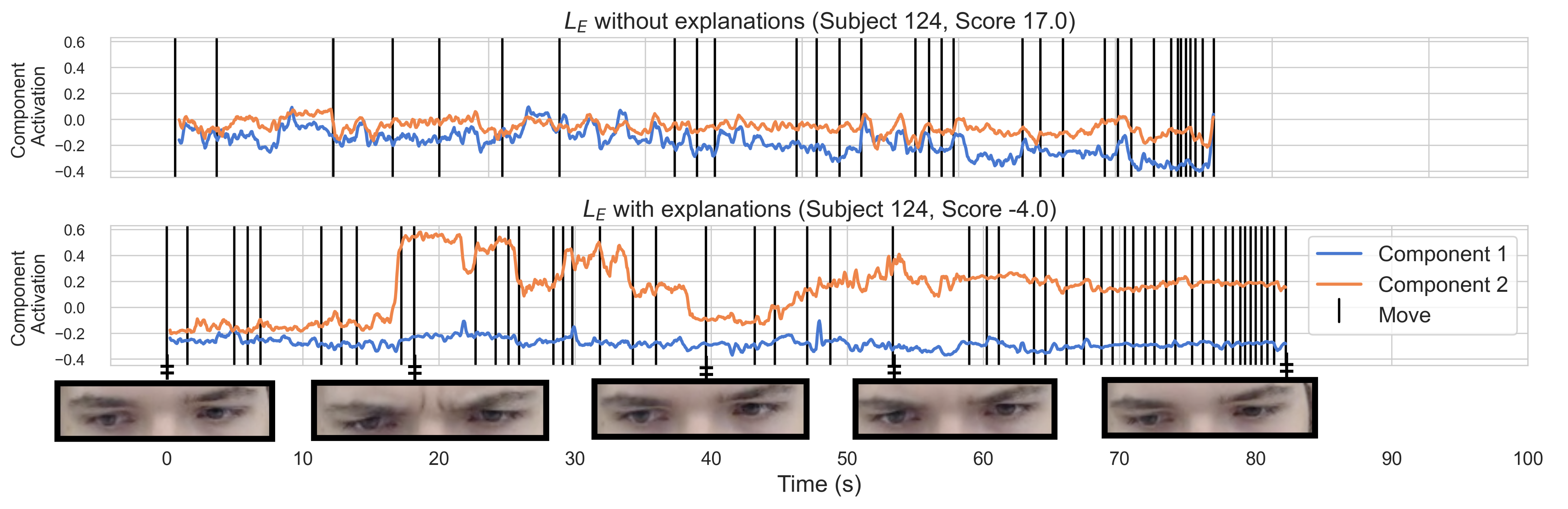}
  \caption{Temporal visualizations of Component 1 and Component 2 as subjects participate in the N and X condition. } \label{fig:activations}
  \vspace*{-5mm}
\end{figure*}



\begin{figure}
  \includegraphics[trim={1mm 1mm 1mm 1mm},clip,width=\linewidth]{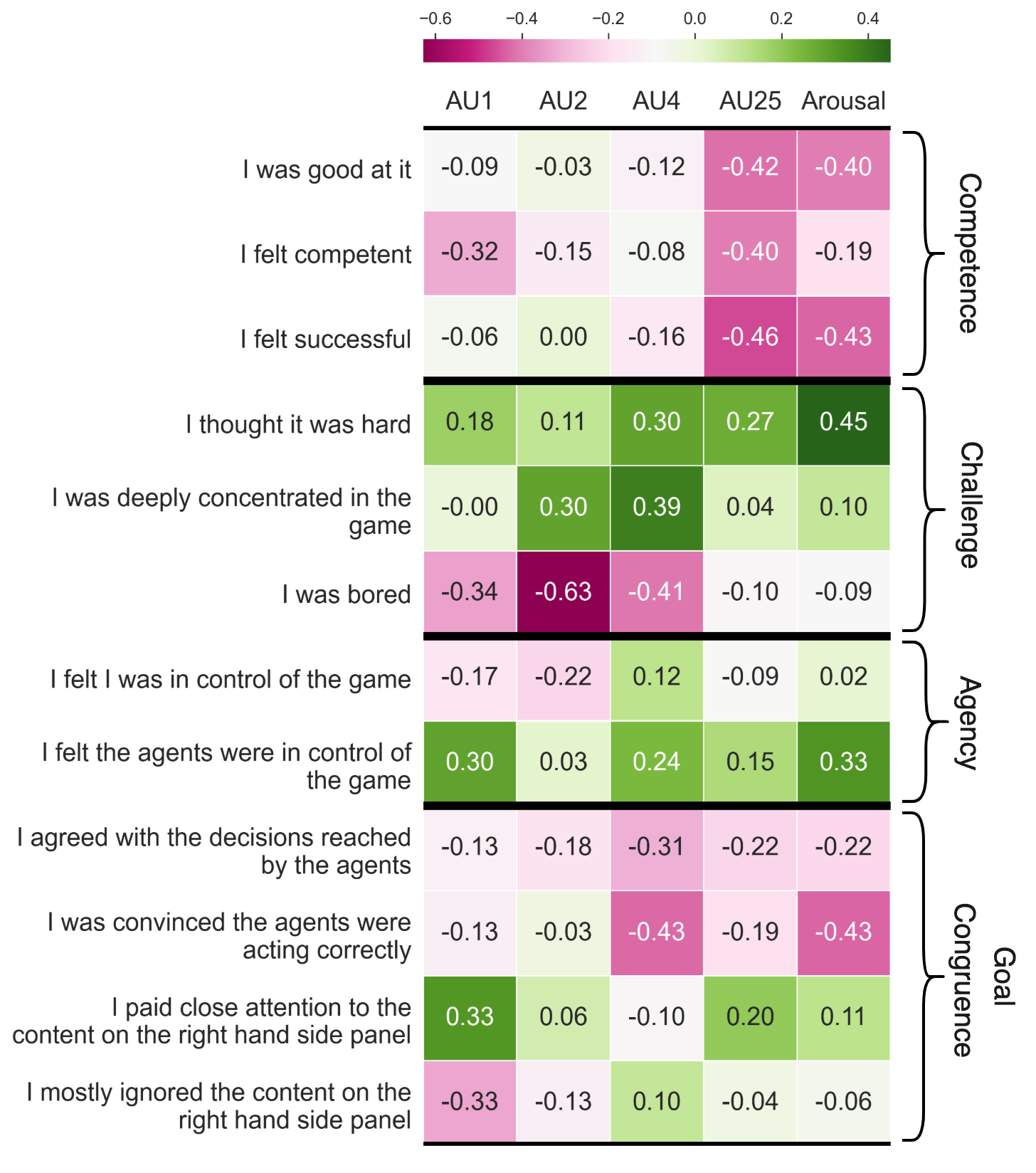}
  \caption{Correlation of change in affect features and change in UX ratings between $X$ and $N$.}\label{fig:ux_affect}
  \vspace*{-5mm}
\end{figure}

\subsection{Component Analysis}

We now examine how a combination of the affect prediction results discussed in Section \ref{sec:hyps} may be combined to form emotion components using PCA. First, we provide quantitative results examining how emotion features are weighted within a component (via PCA \textit{loadings}). Second, we provide qualitative results highlighting how components relate to participant explanation use. 

\subsubsection{Quantitative Results} The top-5 ranked PCA components describe 15\%, 11\%, 8\%, 6\%, and 6\% of the variance in the concatenated multitask feature set. Table \label{table:loadings} shows the 14 loadings with the largest absolute value for Component 1 (15\% of variance) and Component 2 (11\% of variance), where each loading coefficient indicates how much the feature contributes to the component activation. The loadings indicate consistent features used across methods. For example, Component 1 includes AU25 from all three extraction methods, and AU2 / AU1 from the Multitask CNN and OpenFace. These results are interesting because in the hypotheses evaluations provided above, AU1, AU2, and AU25 showed sensitivity between the N and X conditions, and during collision/success events. Component 2 shows a similar result, with AU4 contributing a positive loading across extraction methods and the Neutral emotion category contributing a negative loading across both multitask methods. 

\vspace*{-3mm}
\subsubsection{Qualitative Results} To examine how components describe participant behavior as they interact with explanations, temporal activations of Component 1 and 2 are plotted for subject 124 in Figure \ref{fig:activations} as an illustrative and representative example. 
Mirroring the general trend in \cite{raymond2020culture}, this participant completed $L_{E}$ and demonstrated worse performance with explanations. Both components appear to be responsive to game events, where Component 2 in particular peaks simultaneously with moves.
For example, the top plot of Figure \ref{fig:activations} shows lower Component 2 activation throughout the $N$ condition. However, 18 seconds into the $X$ condition, Component 2 activation peaks and remains elevated thereafter. This event coincides with pronounced brow furrowing (AU4) facial behaviors displayed by the participant. This visual observation is consistent with loadings presented in Table \ref{table:loadings}, which indicate that AU4 has a positive contribution and ``neutral" has a negative contribution to its activation. This finding also supports results in Table \ref{table:hyp3}, which shows that Neutral is expressed less frequently and AU4 more frequently during poor participant performance (collision events). 

\subsection{Subjective User Reports}

Component analysis results support the interpretation that explanations interfered with participant 124 performance in the $X$ condition. In terms of appraisal theory, explanations may provide increased clarity and goal-consistency in $L_{H}$ but may impose additional stress and burden to users in $L_{E}$. However, it is unclear which specific aspect of explanations may have impeded performance. 

Therefore, we now connect \textit{objective} facial affect analysis results with \textit{subjective} reflections of participants. Specifically, if emotional appraisals are connected with explanation efficacy, facial affect predictions should be tied to \textit{self-reported} challenge (perceived difficulty level), competence (perceived ability to succeed), agency (perceived control over the outcome), and goal-congruence (perceived utility of explanations in supporting performance).

We investigate this connection by analyzing participant experience reports collected at the end of each session. Following the protocol outlined in Section \ref{sec:hyps} for affect prediction results, we compute $DUX_u^j$ = $X_u^j - N_u^j$, the difference in each user experience question $u$ for each participant $j$ between the $X$ and $N$ conditions. Figure \ref{fig:ux_affect} shows the Pearson correlation of the change in user experience ratings (DUX) and change in facial affect features (D). 

The top question group shows that as participants feel more competent, AU25 activations (lips parted) are predicted less frequently and estimated arousal is lower. This matches results in Table \ref{table:hyp3}, which show that AU25 increases during low-performance collision events. The second group shows that as participants feel more challenged, AU4 and AU25 are predicted more frequently and estimated arousal is higher. 
As shown in the third group, the frequency of AU1 (inner brow raised), AU4 (frown), and arousal increases when participants feel agents -- rather than themselves -- are in control of the game. Interestingly, these same features are \textit{less frequent} when explanations are provided in $L_H$ (Figure \ref{fig:hyp2}). This suggests that explanations may provide participants with a greater sense of control over the game outcome in difficult conditions. More specifically, as shown in the bottom group, AU4 is predicted less frequently when participants are ``convinced agents were acting correctly" and ``agreed with decisions of the agents". Similarly, AU1 is more frequent when participants ``paid close attention to the content (explanations) on the right hand side of the panel``.

\section{Discussion and Conclusion} \label{section:conclusion}

Our work methodologically shows that ensembled multitask predictions provide robust and interpretable representations of human behavior. Given an affect signal with unknown ground truth, we show how to derive a reliability estimate by examining inter-model prediction agreement. We also show that given multiple emotion representations, componential modeling can be used to examine the importance of each feature (Table \ref{table:loadings}) and its temporal activation (Fig.\ \ref{fig:activations}). This provides a more reliable, interpretable alternative to using a single model in isolation.

Our analyses show that affective signals are correlated with \textit{objective} user performance and \textit{subjective} usability ratings. This opens the door to using real-time affect predictions to improve XAI interfaces. Specifically, in domains such as machine-assisted medical decision making, it is difficult to assess explanation efficacy because performance feedback is sparse \cite{yang2019unremarkable}. In such cases, affective signals may be used to tailor explanations to (1) contextual factors (i.e.\ task difficulty), and (2) user-specific characteristics (i.e.\ perceived competence and agency) in real-time. For example, AU25 and arousal may provide real-time proxies for perceived competence, while AU04 and AU02 could be used to measure perceived challenge. Before real-world deployments, additional research is necessary to identify how these findings generalize to different XAI tasks.\\


\textbf{Acknowledgments:} A. Raymond is supported by L3Harris ASV and the Royal Commission for the Exhibition of 1851. H. Gunes is supported by the EPSRC project ARoEQ under grant ref. EP/R$030782$/$1$. 

{\small
\bibliographystyle{ieee_fullname}
\bibliography{egbib}
}

\end{document}